\documentclass[runningheads]{llncs}

\usepackage{graphicx}
\usepackage[misc]{ifsym} 
\usepackage{bbding}
\usepackage{multirow}
\usepackage{subfigure}
\usepackage{CJKutf8}

\usepackage{floatrow}  
\newfloatcommand{capbtabbox}{table}[][\FBwidth]  
\usepackage{array}
\newcolumntype{I}{!{\vrule width 1.5pt}}

\begin{document}
\title{A3Net:Adversarial-and-Attention Network for Machine Reading Comprehension}
\titlerunning{Adversarial-and-Attention Network for Machine Reading Comprehension}
%
\author{Jiuniu Wang\inst{1,2} \and
	Xingyu Fu\inst{1} \and Guangluan Xu\inst{1(}\Envelope\inst{)} \and
	Yirong Wu\inst{1,2} \and Ziyan Chen\inst{1} \and Yang Wei\inst{1} \and Li Jin\inst{1}}
\authorrunning{J. Wang et al.}
\titlerunning{Adversarial-and-Attention Network for Machine Reading Comprehension}

\institute{Key Laboratory of Technology in Geo-spatial Information Processing and Application System, Institute of Electronics, CAS, Beijing, China\\
	\email{gluanxu@mail.ie.ac.cn}\\ \and
	School of Electronic, Electrical and Communication Engineering, University of Chinese Academy of Sciences, Beijing, China\\
	\email{wangjiuniu16@mails.ucas.ac.cn}}
\maketitle 
\begin{abstract}
In this paper, we introduce Adversarial-and-attention Network (A3Net) for Machine Reading Comprehension. This model extends existing approaches from two perspectives. First, adversarial training is applied to several target variables within the model, rather than only to the inputs or embeddings. We control the norm of adversarial perturbations according to the norm of original target variables, so that we can jointly add perturbations to several target variables during training. As an effective regularization method, adversarial training improves robustness and generalization of our model. Second, we propose a multi-layer attention network utilizing three kinds of high-efficiency attention mechanisms. Multi-layer attention conducts interaction between question and passage within each layer, which contributes to reasonable representation and understanding of the model. Combining these two contributions, we enhance the diversity of dataset and the information extracting ability of the model at the same time. Meanwhile, we construct A3Net for the WebQA dataset. Results show that our model outperforms the state-of-the-art models (improving Fuzzy Score from 73.50\% to 77.0\%).
\keywords{Machine Reading Comprehension  \and Adversarial Training \and Multi-layer Attention.}
\end{abstract}
\section{Introduction}
Machine reading comprehension (MRC) aims to teach machines to better read and comprehend, and answer questions posed on the passages that they have seen \cite{hermann2015teaching}. In this paper, we propose a novel model named Adversarial-and-attention Network (A3Net) for MRC. 

The understanding of neural network is shallow, and it is easy to be disturbed by adversarial examples \cite{jia2017adversarial}. So we adopt  Adversarial training(AT) \cite{goodfellow2014explaining} as a regularization method to improve our model's generality and robustness. Previous works apply adversarial perturbations mainly on input signals \cite{goodfellow2014explaining} or word embeddings  \cite{miyato2016adversarial}, acting as a method to enhance data. While we blend these perturbations into different model layers, especially where question-passage interaction takes place. 

The state-of-the-art models, such as Match-LSTM \cite{wang2016machine}, BIADF \cite{seo2016bidirectional}, SAN \cite{liu2017stochastic}, and ReasoNet \cite{shen2017reasonet}, have been proved to be effective in English MRC datasets including CNN/DailyMail \cite{hermann2015teaching} and SQuAD \cite{rajpurkar2016squad}. They all use attention mechanism and pointer network \cite{vinyals2015pointer} to predict the span answer. However, these models tend to apply attention function on the limited layer. Thus they would ignore some deep-seated information. To solve this problem, we adopt multi-layer attention to extract information at each level. In the low-level, attention weight is highly affected by the similarity of word embedding and lexical structure(e.g. {\itshape affix, part of speech,} etc.), which contains syntactic information. While in the high-level, attention weight reflects the abstract concept correlation between passage and question, which contains semantic information.

To sum up, our contributions can be summarized as follows:
\begin{itemize}
\item We blend adversarial training to each layer of our model. Not only can adversarial training enhance the information representation ability, but also improve extraction ability of the whole model.

\item We apply multi-layer attention to each layer of our model. In this way, our model can make efficient interactions between questions and passages, so as to find which passage span is needed.

\item We propose a novel neural network named A3Net for Machine Reading Comprehension, which gains the best result on the WebQA dataset.
\end{itemize}

\section{Related Work}
{\bfseries Adversarial Training.} Szegedy et al.\cite{szegedy2013intriguing} found that deep neural network might make mistakes when adding small worst-case perturbations to input. This kind of inputs is called adversarial examples. Many models cannot defend the attack of adversarial examples, including state-of-the-art neural networks such as CNN and RNN. In recent years, there are several methods for regularizing the parameters and features of a deep neural network during training. For example, by randomly dropping units, dropout is widely used as a simple way to prevent neural networks from overfitting. 

Adversarial training(AT) \cite{goodfellow2014explaining} is a kind of regularizing learning algorithms. It was first proposed to fine tune the task of image classification. By adding perturbations to input signals during training, neural network gains the ability to tolerant the effect of adversarial example. Miyato et al.\cite{miyato2016adversarial} first adopt AT to text classification. They add perturbations to word embedding and obtain similar benefits like that in image classification. Following this idea, Wu et al.\cite{wu2017adversarial} utilizes AT to Relation Extraction and improves the precision. In order to improve the model stability and generality, we adopt adversarial training to several target variables within our model.

\begin{table}
	\centering
	\caption{An outline of attention mechanism used in state-of-the-art architectures.}\label{tab1}
	\begin{tabular}{|c|c|c|c|}
		\hline
		Model &  Syntactic attention  &  Sematic attention  &  Self-match attention \\
		\hline
		DrQA \cite{chen2017reading} & $\surd$ &  &  \\
		FastQA \cite{weissenborn2017making}& $\surd$ &  &  \\
		Match-LSTM \cite{wang2016machine}&  & $\surd$ &  \\
		BIDAF \cite{seo2016bidirectional}&  & $\surd$ &  \\
		R-Net \cite{Group2017R}&  & $\surd$ & $\surd$ \\
		SAN \cite{liu2017stochastic}&  &  & $\surd$\\
		FusionNet \cite{huang2017fusionnet}& $\surd$ & $\surd$ & $\surd$ \\
		\hline
	\end{tabular}
\end{table}

\noindent {\bfseries Attention mechanism.} Attention mechanism has demonstrated success in a wide range of tasks. Bahdanau et al.  \cite{bahdanau2014neural} first propose attention mechanism and apply it to neural machine translation. And then, it is widely used in many tasks including MRC. Attention near embedding module aims to attend the embedding from the question to the passage \cite{chen2017reading}. Attention after context encoding extracts the high-level representation in the question to augment the context. Self-match attention \cite{tan2017s} takes place before answer module. It dynamically refines the passage representation by looking over the whole passage.

As shown in Table~\ref{tab1}, three different types of attention mechanisms are widely used in state-of-the-art architectures. DrQA \cite{chen2017reading} simply use a bilinear term to compute the attention weights, so as to get word level question-merged passage representation. FastQA \cite{weissenborn2017making} combines feature into the computation of word embedding attention weights. Match-LSTM \cite{wang2016machine} applies LSTM to gain more context information, and concatenate attentions from two directions. A less memory attention mechanism is introduced in BIDAF \cite{seo2016bidirectional} to generate bi-directional attention flow. R-Net \cite{Group2017R} extends self-match attention to refine information over context. SAN \cite{liu2017stochastic} adopts self-match attention and uses stochastic prediction dropout to predict answer during training. Huang et al.\cite{huang2017fusionnet} summarizes previous research and proposes the fully-aware attention to fuse different representations over the whole model. Different from the above models, we utilize three kinds of effective method to calculate attention weights. It helps our model to interchange information frequently, so as to select the appropriate words in passages.

\section{Proposed Model}

\noindent In this paper, we use span extraction method to predict the answer to a specific question $Q = \{ {q_1},{q_2},..,{q_J}\} $ based  on the related passage $P = \{ {p_1},{p_2},...,{p_T}\} $. 

\begin{figure}
	\centering
	\includegraphics[width=12cm , height=8.5cm]{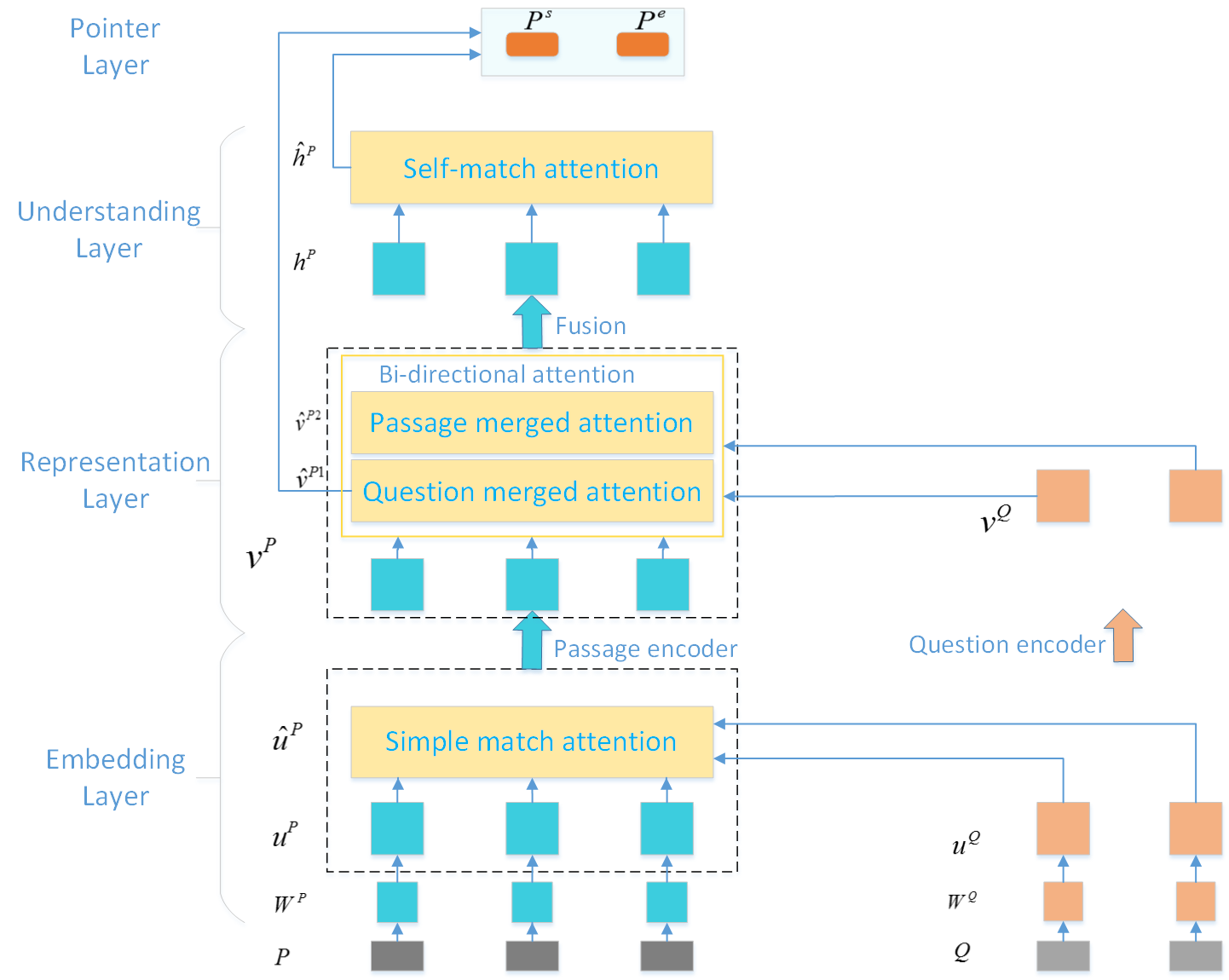}
	\caption{The overall structure of our model. The dotted box represents concatenate operation.} \label{fig1}
\end{figure}

As depicted in Fig.~\ref{fig1}, our model can be decomposed into four layers: Embedding Layer, Representation Layer, Understanding Layer and Pointer Layer. And three attention mechanisms are applied to different layers. Simple match attention is adopted in Embedding Layer, extracting syntactic information between the question and passage. While in Representation Layer, bi-directional attention raises the representation ability by linking and fusing semantic information from the question words and passage words. In Understanding Layer, we adopt self-match attention to refine overall understanding. 

\subsection{Embedding Layer}
\noindent {\bfseries Input Vectors.} We use randomly initialized character embeddings to represent text. Firstly, each word in $P$ and $Q$ is represented as several character indexes. Afterwards, each character is mapped to a high-density vector space(${w^P}$ and ${w^Q}$). In order to get a fixed-size vector for each word, 1D max pooling is used to merge character vectors into word vector(${u^P}$ and ${u^Q}$). Simple match attention is then applied to match word level information, which can be calculated as follows: 
\begin{equation}
\hat u^P = SimAtt(u^P,u^Q)
\end{equation}
where $SimAtt(\cdot)$ denotes the function of simple match attention.

\noindent {\bfseries Simple Match Attention.} Given two sets of vector ${v^A} = \{ v_1^A,v_2^A,...,v_N^A\} $ and ${v^B} = \{ v_1^B,v_2^B,...,v_M^B\} $, we can synthesize information in ${v^B}$ for each vector in ${v^A}$. Firstly we get the attention weight $\alpha _{ij}$ of $i$-th word in $A$ and $j$-th word in $B$ by $\alpha _{ij} =softmax(\exp (<v_i^A,v_j^B>))$, where $<>$ represents inner product. Then calculate the sum for every vector in $v^B$ weighted by $\alpha _{ij}$ to get the attention representation $\hat v_i^A = \sum\limits_j {\alpha _{ij}v_j^B}$. So attention variable $\hat v^A$ can be denoted as $\hat v^A = \{ \hat v_1^A,\hat v_2^A,...,\hat v_N^A\}  = SimAtt(v^A,v^B)$.

\subsection{Representation Layer}

To better extract semantic information, we utilize RNN encoders to produce high-level representation $v_1^Q,...,v_J^Q$ and $v_1^P,...,v_T^P$ for all words in the question and passage respectively. The encoders are made up of bi-directional Simple Recurrent Unit (SRU) \cite{lei2017training}, which can be denoted as follows:
\begin{equation}
v_t^P = BiSRU(v_{t - 1}^P,[u_t^P;\hat u_t^P])
,
v_j^Q = BiSRU(v_{j - 1}^Q,u_j^Q)
\end{equation}

\noindent {\bfseries Bi-directional attention} is applied in this layer to combine the semantic information between questions and passages. Similar to the attention flow layer in BIDAF, we compute question merged attention ${\hat v^{P1}}$ and passage merged attention ${\hat v^{P2}}$  with bi-directional attention. The similarity matrix is computed by ${S_{ij}} = \beta (v_i^P,v_j^Q)$, we choose
\begin{equation}
\beta (v_i^P,v_j^Q) = {W_{(S)}}^T[v_i^P;v_j^Q;v_i^P \cdot v_j^Q]
\end{equation}

\noindent where ${W_{(S)}}$ is trainable parameters, $ \cdot $ is element-wise multiplication, $[;]$ is vector concatenation across row.

{\itshape Question merged attention} signifies which question words are most relevant to each passage words. Question merged attention weight(the $i$-th word in passage versus a certain word in question) is computed by ${a_{i:}} = {\rm{softmax(}}{{\rm{S}}_{i:}}{\rm{)}} \in {{\rm{R}}^J}$. Subsequently, each attended question merged vector is denoted as ${\hat v_i}^{P1} = \sum\nolimits_j {{a_{ij}}v_j^Q} $.
{\itshape Passage merged attention} signifies which context words have the closest similarity to one of the question words and hence critical for answering the question. The attended passage-merged vector is ${\tilde v^{P2}} = \sum\nolimits_i {{b_i}v_i^P} $, where $b = {softmax}(ma{x_{col}}(S))$ and $b \in {R^T}$, the maximum function $ma{x_{col}}()$ is performed across the column. Then ${\tilde v^{P2}}$ is tiled T times to ${\hat v^{P2}} \in {R^{2d \times T}}$, where $d$ is the length of hidden vectors.

\subsection{Understanding layer}
The above bi-directional attention representation ${\hat v_i}^{P1}$ and ${\hat v^{P2}}$ is concatenated with word representation $v^p$ to generate the attention representation ${\hat v^P}$.
\begin{equation}
{\hat v^P} = [{\hat v^{P1}};{\hat v^{P2}};{v^P}]
\end{equation}

Then we use a bi-directional SRU as a Fusion to fuse information, which can be represented as $h_t^P = BiSRU(h_{t - 1}^{P - 1},{\hat v_t}^P)$.

In order to take more attention over the whole passage, we apply self-match attention in Understanding Layer. Note that the computing function is the same as simple match attention, but its two inputs are both ${h^P}$
\begin{equation}
\hat h^P = SimAtt(h^P,h^P)
\end{equation}

\subsection{Pointer Layer}
Pointer network is a sequence-to-sequence model proposed by Vinyals et al.\cite{vinyals2015pointer} In Pointer Layer, we adopt pointer network to calculate the possibility of being the start or end position for every word in the passage. Instead of using a bilinear function, we take a linear function (which is proved to be simple and effective) to get the probability of start position ${P^s}$ and end position ${P^e}$
\begin{equation}
{P^s} = {softmax}({W_{Ps}}[\hat h_i^P;\hat v_i^{P1}]), {P^e} = {softmax}({W_{Pe}}[\hat h_i^P;\hat v_i^{P1}; P^s])
\end{equation}

\noindent {\bfseries Training.} During training, we minimize the cross entropy of the golden span start and end $L(\theta ) = \frac{1}{N}\sum\limits_k^N {(\log (P_{i_k^s}^s) + \log (P_{i_k^e}^e))} $, where $i_k^s$, $i_k^e$ are the predicted answer span for the k-th instance.

\noindent {\bfseries Prediction.} We predict the answer span to be $i_k^s$, $i_k^e$ with the maximum $P_{{i^s}}^s + P_{{i^e}}^e$ under the constraint $0 \le {i^e} - {i^s} \le 10$.

\subsection{Adversarial training}
\begin{figure}[tb]
	\vspace{-0.8cm}  
	\centering
	\includegraphics[width=6cm, height=5.2cm]{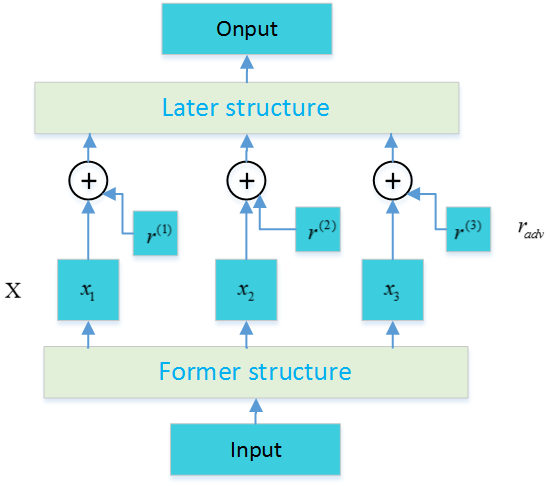}
	\caption{The computation graph of adversarial training. $X$  denotes target variable, ${r_{adv}}$ denotes adversarial perturbation. The Input of the model is mapped into target variable $X$ by Former Structure. And then Later Structure generates the Output based on the target variable $X$ conbimed with adversarial perturbation ${r_{adv}}$.} \label{fig2}
\end{figure}

Adversarial training applies worst case perturbations on target variables. As is shown in Fig.~\ref{fig2}, we denote $X$ as the target variable and $\theta $ as the parameters of the model. Different from previous works, $X$ can be set as each variable in our model, adversarial training adds adversarial cost function ${L_{adv}}(X;\theta )$ to the original cost function. 
The equation of ${L_{adv}}(X;\theta )$ is described as follows:
\begin{equation}
{L_{adv}}(X;\theta ) = L(X + {r_{adv}};\theta ) , {r_{adv}} = \arg \mathop {\max }\limits_{||r|| \le \varepsilon ||X|| } L(X + r;\hat \theta )
\label{e9} 
\end{equation}

\noindent where $r$ is a perturbation on the target variable and $\hat \theta $ is a fixed copy to the current parameters. When optimizing parameters, the gradients should not propagate through ${r_{adv}}$. One problem is that we cannot get the exact value of ${r_{adv}}$ simply following Eqs.(\ref{e9}), since the computation is intractable. Goodfellow et al.\cite{miyato2016adversarial} approximate the value of ${r_{adv}}$ by linearizing $L(X;\hat \theta )$ near $X$

\begin{equation}
{r_{adv}} = \varepsilon ||X||\frac{g}{{||g||}} , g = {\nabla _X}L(X|\hat \theta )
\label{e10} 
\end{equation}

\noindent where $|| \cdot ||$ denotes the norm of variable $ \cdot $  , and $\varepsilon $ is an intensity constant to adjust the relative norm between $||{r_{adv}}||$  and $||X||$. So the norm of ${r_{adv}}$ is decided by $||X||$, and it could be different for each training sample and training step.

\section{Experiments}
In this section, we evaluate our model on the WebQA dataset. Outperforming the baseline model in the original paper (Fussy Score 73.50\%), we obtain Fussy Score 77.01\% with multi-layer attention and adversarial training. 

\subsection{Dataset and evaluation metrics}

\begin{table*}[b]
	\begin{floatrow}  
		\capbtabbox{  
			\begin{tabular}{|p{4.5cm}|}
				\hline
				{\bfseries Question:}\\ What kind of color can absorb all the seven colors of the sunshine?
				(\begin{CJK}{UTF8}{gbsn}哪种颜色的物体能把太阳的七种颜色全部吸收？\end{CJK}) \\
				{\bfseries Passage:}\\ Black absorbs light of various colors
				(\begin{CJK}{UTF8}{gbsn}黑色能够吸收各种颜色的光\end{CJK})\\
				{\bfseries Answer:} Black(\begin{CJK}{UTF8}{gbsn}黑色\end{CJK})\\
				\hline
			\end{tabular}
		}{  
			\caption{An example from WebQA.}  
			\label{tab:tab0}  
		}  
		\capbtabbox{  
			\begin{tabular}{|c|r|r|r|r|}
				\hline
				\multirow{2}*{Dataset} & \multicolumn{2}{c|}{question}  & \multicolumn{2}{c|}{annotated evidence} \\
				\cline{2-5}
				& \multicolumn{1}{c|}{\#} & \multicolumn{1}{c|}{word\#} & \multicolumn{1}{c|}{\#} & \multicolumn{1}{c|}{word\#} \\
				\hline
				Train & 36,145 & 374,500 & 140,897 & 10,757,652 \\
				Validation & 3,018 & 36,666 & 5,412 & 233,911 \\
				Test & 3,024 & 36,815 & 5,445 & 234,258\\
				\hline
			\end{tabular}
		}{  
			\caption{Statistics of WebQA dataset.}  
			\label{tab:tab2}  
		}
	\end{floatrow}  
\end{table*}  

\noindent WebQA \cite{li2016dataset} is a large scale real-world Chinese QA dataset. Table~\ref{tab:tab0} gives an example from WebQA dataset. Its questions are from user queries in search engines and its passages are from web pages. Different from SQuAD, question-passage pairs in WebQA are matched more weakly. We use annotated evidence(shown in Table~\ref{tab:tab2}) to train and evaluate our model. There is an annotated golden answer to each question. So we can measure model performance by comparing predicted answers with golden answers. It can be evaluated by precision (P), recall (Q) and F1-measure (F1): 

\begin{equation}
P = \frac{{|C|}}{{|A|}},R = \frac{{|C|}}{{|Q|}},F1 = \frac{{2PR}}{{P + R}}
\end{equation}

\noindent where $|C|$ is the number of correctly answered questions, $|A|$ is the number of produced answers, and $|Q|$ is the number of all questions. 

The same answer in WebQA may have different surface forms, such as “Beijing” v.s. “Beijing city”. So we use two ways to count correctly answered questions, which are referred to as {\itshape Strict} and {\itshape Fuzzy}. Strict matching means the predicted answer is identical to the standard answer; Fuzzy matching means the predicted answer can be a synonym of the standard answer.

\subsection{Model details}
In our model, we use randomly initialized 64-dimensional character embedding and hidden vector length $d$ is set to 100 for all layers. We utilize 4-layer Passage encoder and Question encoder. And Fusion SRU is set to 2-layer. We also apply dropout between layers, with a dropout rate of 0.2. The model is optimized using AdaDelta with a minibatch size of 64 and an initial learning rate of 0.1. Hyper-parameter $\varepsilon $ is selected on the WebQA validation dataset.

\subsection{Main results}
\begin{table}[b]
	\caption{Evaluation results on the test dataset of WebQA.}\label{tab3}
	\centering
	\begin{tabular}{|c|c|c|c|c|c|c|}
		\hline
		\multirow{2}*{Model} & \multicolumn{3}{c|}{Strict Score}  & \multicolumn{3}{c|}{Fuzzy Score} \\
		\cline{2-7}
		& P(\%) & R(\%) & F1(\%) & P(\%) & R(\%) & F1(\%) \\
		\hline
		LSTM+softmax & 59.38 & 68.77 & 63.73 & 63.58 & 73.63 & 68.24 \\
		LSTM+CRF & 63.72 & \textbf{76.09} & 69.36 & 67.53 & \textbf{80.63} & 73.50 \\
		BIDAF & 70.04 & 70.04 & 70.04 & 74.43 & 74.43 & 74.43\\
		A3Net base model(without AT) & 71.03 & 71.03 & 71.03 & 75.46 & 75.46 & 75.46 \\
		A3Net(random noise) & 71.28 & 71.28 & 71.28 & 75.89 & 75.89 & 75.89\\
		A3Net & {\bfseries 72.51} & 72.51 & {\bfseries 72.51} & {\bfseries 77.01} & 77.01 & {\bfseries 77.01}\\
		\hline
	\end{tabular}
\end{table}

The evaluation results are shown in Table~\ref{tab3}. Different models are evaluated on the WebQA test dataset, including baseline models(LSTM+softmax and LSTM+CRF), BIDAF and A3Net. A3Net base model(without AT) denotes our multi-layer attention model which does not apply adversarial training(AT); A3Net(random noise) denotes control experiment which replaces adversarial perturbations with random Gaussian noise with a scaled norm. Baseline models utilize sequence label method to mark the answer, while others adopt pointer network to extract the answer. Sequence label method can mark several answers for one question, leading high recall(R) but low precision(P). So we adopt pointer network to generate one answer for each question. In this condition, evaluation metrics(P, R, F1) are equal. Thus we can use {\itshape Score} to evaluate our model. Besides, Fuzzy evaluation is closer to real life, so we mainly focus on {\itshape Fuzzy Score}.

Based on single layer attention and pointer network, BIDAF obtains the obvious promotion (Fuzzy F1 74.43\%). Benefit from multi-layer attention, A3Net base model gains 0.97\% promotion in Fuzzy Score compared to BIDAF. It indicates that multi-layer attention is useful. Our model would get another 1.12\% promotion in Fuzzy Score after jointly adopting adversarial training on target variable ${w^P}$ and ${\hat v^P}$. 

A common misconception is that perturbation in adversarial training is equivalent to random noise. In actually, noise is a far weaker regularization than adversarial perturbations. An average noise vector is approximately orthogonal to the cost gradient in high dimensional input spaces. While adversarial perturbations are explicitly chosen to consistently increase the loss value. To demonstrate the superiority of adversarial training over the addition of noise, we include control experiments which replaced adversarial perturbations with random perturbations from a Gaussian distribution. We use random noise to replace worst case perturbations on each target variable, which only lead slightly improvement. It indicates that AT can actually improve the robustness and generalization of our model.

\subsection{Ablation on base model structure}

\begin{table}
	\caption{Comparison of different configurations of base model. The symbols in this table is corresponding with Fig.~\ref{fig1}.} 
	\label{tab4}
	\centering
	\begin{tabular}{|c|c|c|}
		\hline
		model & Strict Score(\%) & Fuzzy Score(\%) \\
		\hline
		A3Net base model (without ${\hat u^P}$)& 70.57 & 74.93  \\
		A3Net base model (without ${\hat v^{P1}}$) & 70.77&75.18  \\
		A3Net base model (without ${\hat v^{P1}}$ and ${\hat v^{P2}}$) & 70.63& 74.56 \\
		A3Net base model (without ${\hat h^P}$) & 70.70& 75.23\\
		A3Net base model & {\bfseries 71.03} & {\bfseries 75.46}\\
		\hline
	\end{tabular}
\end{table}

Next, we investigate the ablation study on the structure of our base model. From Table~\ref{tab4}, we can tell that both Strict Score and Fuzzy Score would drop when we omit any designed attention. It indicates that each attention layer in A3Net base model is essential.

\subsection{Adversarial training on different target variables}

\begin{table}[b]
	\caption{Comparison of adversarial training results on different target variables. The symbols in this table is corresponding with Fig.~\ref{fig1}}\label{tab5}
	\centering
	\begin{tabular}{|c|c|cIc|c|c|}
		\hline
		Target variable & Strict Score & Fuzzy Score & Target variable & Strict Score & Fuzzy Score \\
		\hline
		none (base model) & 71.03 & 75.46 & ${\hat v^{P1}}$ & 71.85 & 76.28 \\
		${w^P}$& 71.95 & 76.62 & ${\hat h^{P}}$ & 71.56 & 76.42 \\
		${u^P}$ & 72.06 & 76.39& ${\hat v^{P}}$ & 72.28 & 76.81 \\
		${\hat u^P}$& 71.32 & 75.92 & ${w^P}$ and ${\hat v^P}$ & {\bfseries 72.51} &  {\bfseries 77.01} \\
		\hline
	\end{tabular}
\end{table}

\noindent We evaluate the predicted result when we apply adversarial training on different target variables. As is shown in Table~\ref{tab5}, applying adversarial training on each target variable can improve Fuzzy Score as well as Strict Score in different degree. It indicates that adversarial training can work as a regularizing method not only for word embeddings, but also for many other variables in our model. Note that the Score is improved significantly when applying AT on embedding variable ${w^P}$ and attention variable ${\hat v^P}$. It reveals that adversarial training can improve representing ability for both inputs and non-input variables. Finally, we obtain the best result when applying AT on both ${w^P}$ and ${\hat v^P}$.

\begin{figure}[b]
	\centering
	\includegraphics[width=8cm, height=5cm]{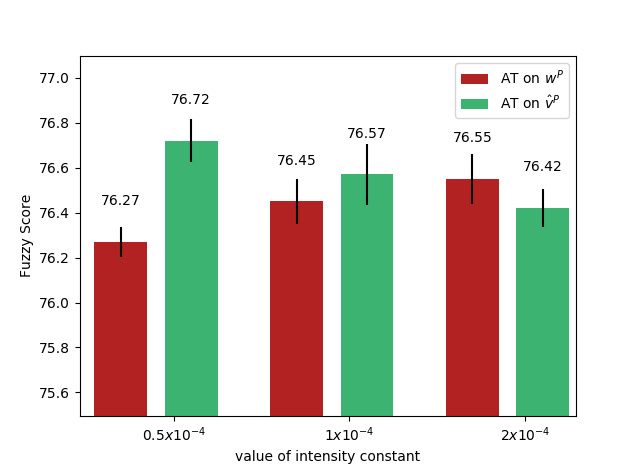}
	\caption{Effect of intensity constant when AT on target variable ${w^P}$ and ${\hat v^P}$.} \label{fig3}
\end{figure}

We also evaluate adversarial training on two target variables (${w^P}$ and ${\hat v^P}$) under different intensity constant $\varepsilon $. As  shown in Fig.~\ref{fig3}, we repeat experiment 3 times for each target variable on each constant $\varepsilon $, and get the average Fuzzy Score and its std. error. For AT on attention variable ${\hat v^P}$, we obtain the best performance when $\varepsilon $ is $0.5 \times {10^{ - 4}}$; While for AT on character embedding variable ${w^P}$, we obtain the best performance when $\varepsilon $ is $2 \times {10^{ - 4}}$. It indicates we need larger adversarial perturbation for low-level variable. We can explain this phenomenon in two different views. Firstly, ${w^P}$ and ${\hat v^P}$ are in different concept levels. ${w^P}$ contains syntactic meaning, and represents as character embedding vectors. Most of the vectors can still hold original meaning under small perturbation, because most points in embedding space have no real meanings. But ${\hat v^P}$ contains semantic meaning. Any perturbation on it would change its meaning, thus our model is sensitive to the perturbation on ${\hat v^P}$. Secondly, ${w^P}$ and ${\hat v^P}$ are in different layers of our model. ${\hat v^P}$ is closer to Pointer Layer, which could have more influence on the output of the model and computation of cost function. 

\subsection{Effective of adversarial training}

\begin{figure}
	\centering
	\subfigure[Fuzzy Score (test) under different training step.]{\label{fig4:a}
	\includegraphics[width=0.7\linewidth]{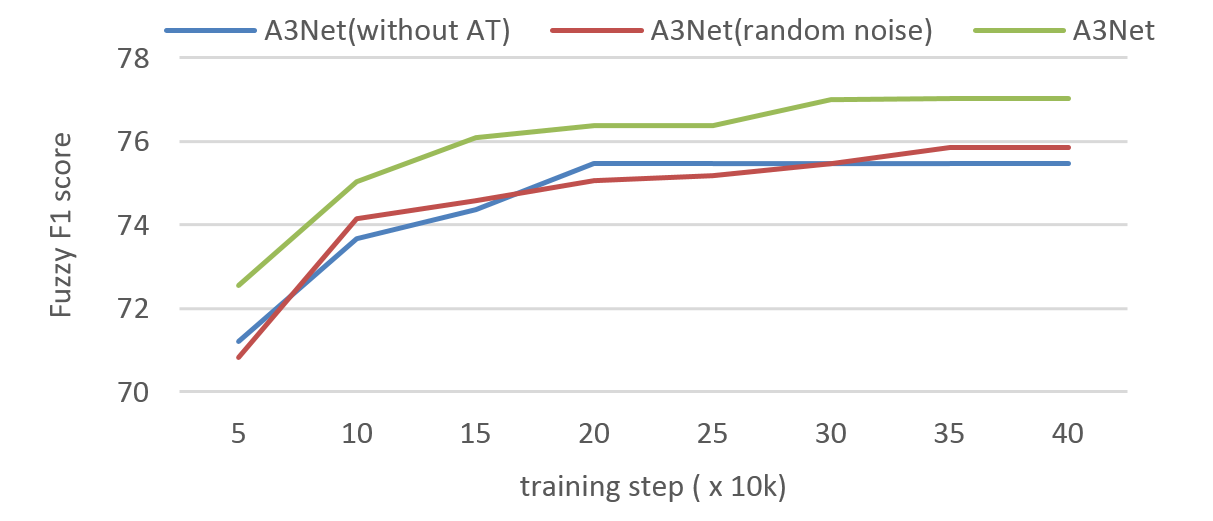}}

	\subfigure[Loss value (train) under different training step.]{\label{fig4:b}
	\includegraphics[width=0.7\linewidth]{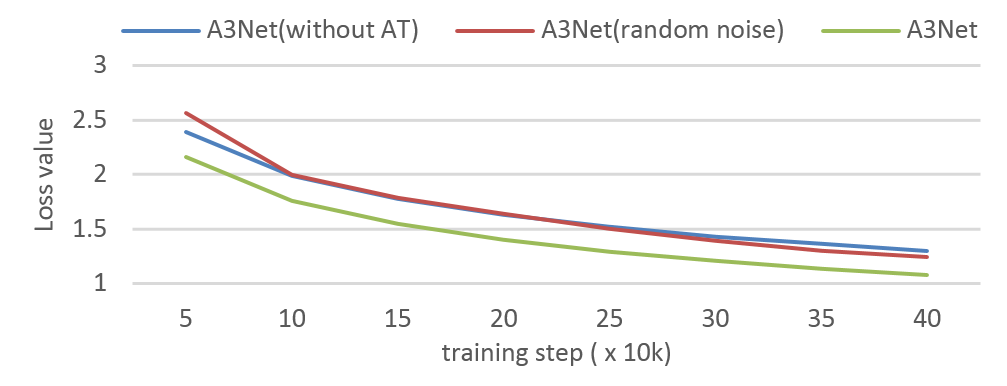}}
	\caption{Fuzzy Score (test) and Loss value (train) under different training step.}
	\label{fig4}
\end{figure}

\noindent Fig.~\ref{fig4:a} shows the Fuzzy Score on the test dataset and Fig.~\ref{fig4:b} shows the loss value on the training dataset of A3Net under different configurations. The data curves of the base model and the random noise model are close to each other in both two figures. It indicates that random noise has limited effect on our model. Within each training step, the Fuzzy Score of our final model is the highest, and its loss value is the lowest. It demonstrates that adversarial training can lead to better performance with less training step.

\section{Conclusions}
This paper proposes a novel model called Adversarial-and-attention Network (A3Net), which includes adversarial training and multi-layer attention.

Adversarial training works as a regularization method. It can be applied to almost every variable in the model. We blend adversarial training into each layer of the model by controlling the relative intensity of norm between adversarial perturbations and original variables. Results show that applying adversarial perturbations on some high-level variables could lead even better performance than that on input signals. Our model obtains the best performance by jointly applying adversarial training to character embedding and high-level attention representation. 

We use simple match attention and bi-directional attention to enhance the interaction between questions and passages. Simple match attention on Embedding Layer refines syntactic information. Bi-directional attention on Representation Layer refines semantic information. Furthermore, self-much attention is used on Understanding Layer to refine the overall information among the whole passages. Experiments on the WebQA dataset show that A3Net outperforms the state-of-the-art models.

\bibliographystyle{splncs04}
\bibliography{mybibtex}

\end{document}